%% file: iclr2025_conference.tex
\documentclass{article} 
\usepackage{iclr2025_conference,times}

\input{math_commands.tex}

\usepackage{hyperref}
\usepackage{url}
\usepackage{graphicx}
\usepackage[most]{tcolorbox} 
\usepackage{caption}
\usepackage{tabularray}
\usepackage{marvosym}

\title{Extract, Match, and Score: An Evaluation Paradigm for Long Question-context-answer Triplets in Financial Analysis}

\author{
Bo Hu\textsuperscript{†} \& Han Yuan\textsuperscript{†} \& Vlad Pandelea\textsuperscript{†} \& Wuqiong Luo\textsuperscript{†} \& Yingzhu Zhao\textsuperscript{†} \& Zheng Ma\textsuperscript{\Letter} \\
Global Decision Science, American Express \\
\makebox[0pt][l]{\texttt{\{Bo.Hu, Han.Yuan1, Vlad.A.Pandelea, Wuqiong.Luo,Yingzhu.Zhao,}}\\
\texttt{Zheng.Ma2\}@aexp.com}
}

\newcommand{\ems}{EMS }
\iclrfinalcopy
\begin{document}

\maketitle
\begin{abstract}
The rapid advancement of large language models (LLMs) has sparked widespread adoption across diverse applications, making robust evaluation frameworks crucial for assessing their performance. While conventional evaluation metrics remain applicable for shorter texts, their efficacy diminishes when evaluating the quality of long-form answers. This limitation is particularly critical in real-world scenarios involving extended questions, extensive context, and long-form answers, such as financial analysis or regulatory compliance. In this paper, we use a practical financial use case to illustrate applications that handle ``long question-context-answer triplets". We construct a real-world financial dataset comprising long triplets and demonstrate the inadequacies of traditional metrics. To address this, we propose an effective Extract, Match, and Score (EMS) evaluation approach tailored to the complexities of long-form LLMs' outputs, providing practitioners with a reliable methodology for assessing LLMs' performance in complex real-world scenarios.
\end{abstract}
\footnotetext{\textsuperscript{†} These authors contributed equally to this work.} \footnotetext{\textsuperscript{\Letter} Correspondence: Zheng Ma, Singapore Decision Science Center of Excellence, American Express, 1 Marina Boulevard, 018989, Singapore.}

\section{Introduction}
The advancement of large language models (LLMs) has revolutionized automated model-based solutions to an unprecedentedly accuracy that rivals human performance across a series of natural language processing (NLP) tasks \citep{10.1145/3649506}. Despite their impressive performance on short-form tasks \citep{10.1145/3641289}, the applicability and effectiveness of LLMs in long-form scenarios warrant in-depth exploration, particularly those where short-form summaries lead to substantial information loss \citep{shaham-etal-2023-zeroscrolls,wijesiriwardene-etal-2023-analogical}. Preliminary studies have investigated long-form challenges in the general domain using either standalone LLMs \citep{han-etal-2023-prewome,liu-etal-2024-longgenbench,li-etal-2024-retrieval,wang-etal-2024-revisiting,laban-etal-2024-summary} or retrieval-augmented generation (RAG) systems \citep{su-etal-2022-read,yu-etal-2023-towards,zhao-etal-2024-longrag,qi-etal-2024-long2rag,han-etal-2024-rag,laban-etal-2024-summary}. 

Compared to the general domain, long-form scenarios are particularly prevalent in the financial sector, where documents such as financial statements and compliance reports often exceed hundreds of pages \citep{liu-etal-2022-long,reddy-etal-2024-docfinqa,zhao-etal-2024-findver}. However, the absence of effective evaluation metrics poses a significant challenge in financial NLP. In particular, traditional metrics lose their efficacy as evaluation tasks increase in length and complexity. Consequently, many studies in financial NLP simplify complex, open-ended analytical tasks into well-defined, closed-ended tasks such as news sentiment analysis \citep{guo-etal-2023-chatgpt,chen-etal-2024-efsa,kirtac-germano-2024-enhanced,shah-etal-2023-trillion}, concept classification \citep{fonseca-cohen-2024-large}, name entity recognition \citep{shah-etal-2022-flue,guo-etal-2023-chatgpt,bhatia-etal-2024-fintral}, structure boundary detection \citep{shah-etal-2022-flue}, event extraction \citep{zhou-etal-2021-trade}, relation identification \citep{li-etal-2023-chatgpt}, etc. While some datasets focus on open-ended tasks such as question answering \citep{shah-etal-2022-flue,reddy-etal-2024-docfinqa,krumdick-etal-2024-bizbench,chen-etal-2024-fintextqa} and abstractive summary \citep{zhu-etal-2024-benchmarking}, they typically address only one or two of the three key aspects that possess the long-form property: context \citep{reddy-etal-2024-docfinqa,srivastava-etal-2024-evaluating,zhu-etal-2024-benchmarking,NEURIPS2023_6a386d70,zhao-etal-2024-findver,gupta-etal-2024-systematic}, question \citep{krumdick-etal-2024-bizbench,chen-etal-2022-convfinqa}, and answer \citep{pmlr-v235-band24a,qi-etal-2024-long2rag}. To the best of our knowledge, while a few efforts have been made in the general domain \citep{tan-etal-2024-proxyqa}, no existing work directly addresses the evaluation challenge of long triplets in finance.

To systematically investigate the current landscape and advance the research on financial long triplets, we (1) Benchmarked the state-of-the-art LLMs on long-form analysis of earning call transcripts from 10 largest constituents in S\&P 500 index; (2) Identified the inefficiency of conventional evaluation methods, such as ROUGE, in differentiating generation quality within the context of long triplets \citep{xu-etal-2023-critical,tan-etal-2024-proxyqa}; (3) Proposed a generalizable evaluation paradigm named EMS (\underline{E}xtract, \underline{M}atch, and \underline{S}core) to provide fine-grained evaluations and demonstrated its competitive advantages over RAGChecker \citep{ru2024ragchecker} in real-world scenarios of long triplets.

\section{Methodology}
To address the inefficiency of conventional evaluation approaches in the context of long triplets, we propose a three-stage methodology \ems: (1) Extracts salient points from both the reference and candidate answers; (2) Identifies potential alignments between these points; (3) Scores each matched pair based on predefined criteria.
The long-form generated answer is denoted as \(ans\), and the corresponding reference is denoted as \(ref\). Our evaluation pipeline takes \(ans\) and \(ref\) as input, and provides three saliency points level metrics: EMS-Recall, EMS-Precision, EMS-F1.

\subsection{Stage 1: Saliency Point Extraction} \label{sec_extracting}

Given a long-form reference answer, we define a function \( f^{(\text{E})}(\cdot) \) that extracts a list of saliency points from the text. Formally,
\begin{equation}
    P^{(\text{ref})} = f^{(\text{E})}(ref) = [p_1^{(\text{ref})},\dots,p_N^{(\text{ref})}],
    \label{equ_extract_ref}
\end{equation}

where \( P^{(\text{ref})} \) denotes the set of saliency points extracted from the reference. We apply the same function \( f(\cdot) \) to the candidate answer:

\begin{equation}
P^{(\text{ans})} = f^{(\text{E})}(ans) = [p_1^{(\text{ans})},\dots,p_M^{(\text{ans})}],
\label{equ_extract_asn}
\end{equation}

yielding the set \( P^{(\text{ans})} \), which contains the saliency points extracted from the candidate answer. 

The purpose of saliency points extraction is to decompose the long-form answer into key claims, ensuring that the evaluation score of these claims accurately reflects the quality of the original response. To achieve this, the saliency point extractor should follow well-crafted instructions.

\begin{figure}[t]
    \centering
    \includegraphics[width=0.75\textwidth]{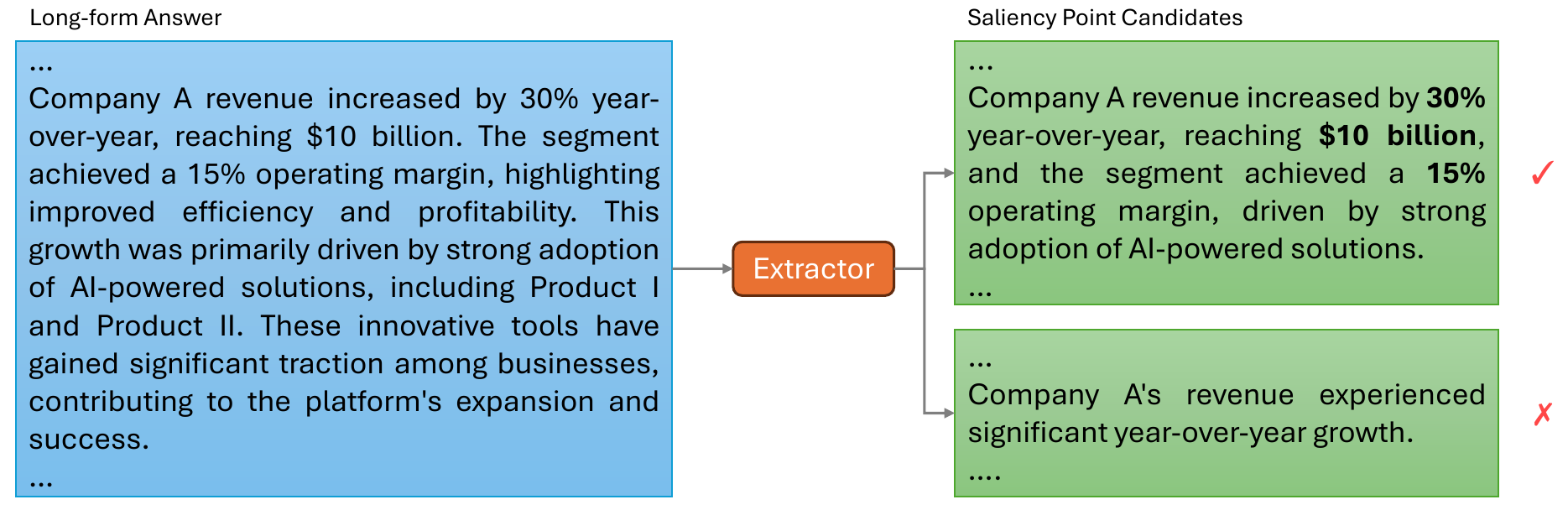}
    \caption{An example of saliency point extraction from long-form answer.}
    \label{extractor_example}
\end{figure}

\textbf{Including details.} Saliency points are not brief summaries from the original answer, since the detailed information is useful to users and shows as a sign of good quality. Hence, saliency points should maintain all the details from the original long-form answer, so that the scoring over saliency points can represent the quality of original answers. Figure \ref{extractor_example} shows an example that the detailed numbers related to the revenue is critical to users with finance background. The shorter extracted point is factually accurate, however, it filters out indispensable information.

\textbf{Retaining repeated points.} Low-quality answers are not solely the result of missing or incorrect claims, which can be penalized during scoring, but also stem from poor organization of those claims. Specifically, a long-form answer should be assigned with a low score if it contains repeated content, even when the saliency points from those repetitions are accurate. Therefore, all repeated points are retained during extraction, leading to a saliency point list that includes all duplicates.

\textbf{Removing summary.} An exception to the treatment of repeated content is with the summaries from the long-form answers. High-quality long-form answers typically include a structured response with an introduction, detailed body, and conclusion. Since the content in the introduction and conclusion often overlaps with the body, including it as duplicated points would be unnecessary. To address this, we exclude summaries, such as overviews and conclusions, before extracting saliency points.

The extractor can be implemented with several solutions, \textit{e.g.}, fuzzy search for summary removal and retaining all details in the saliency points by directly splitting the long-form answer into sentences. This approach works well for answers where the middle section is already well-organized with bullet points. However, it struggles with answers that contain extensive details and examples. To address this, we utilize an LLM-based pipeline to remove summaries from the beginning and the end, then extract saliency points following the specified instructions discussed above. To control the level of detail in the saliency points, we employ in-context learning (ICL) with a few-shot example setup. The prompt used for saliency point extraction is available in Appendix~\ref{sec:extract_prompt}.

\subsection{Stage 2: Matching} \label{sec_matching}
In the second stage of \ems, we apply a function \( f^{(\text{M})} \) to establish correspondences between points in \( P^{(\text{ref})} \) and \( P^{(\text{ans})} \). The output of the matching stage is 
\begin{equation}
    A = [a_1, a_2, \dots, a_N],
    \label{equ_match_list}
\end{equation}

where each index \(a_i\in\{-1,1,2,\dots,M\}\), is computed with the \(i\)-th reference saliency point and the candidate answer saliency point list by:
\begin{equation}
    a_i = f^{(\text{M})}\left( p^{(\text{ref})}_i, P^{(\text{ans})} \right).
    \label{equ_matching}
\end{equation}

\begin{figure}[t]
    \centering
    \includegraphics[width=0.75\textwidth]{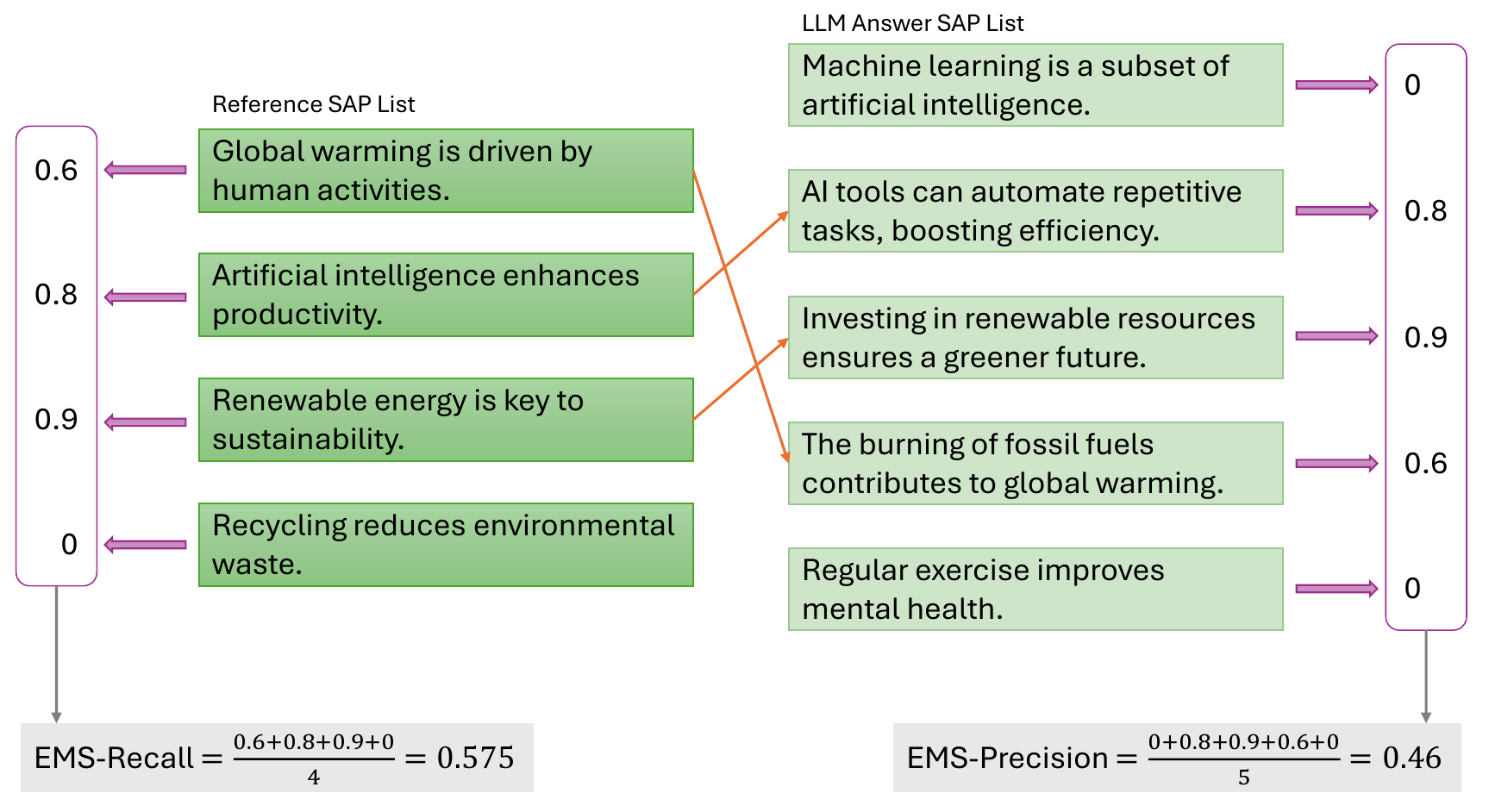}
    \caption{Illustration of matching and scoring procedure in \ems evaluation pipeline. Then EMS-Recall and EMS-Precision are computed by aggregating saliency point level scores.}
    \label{matcher_scorer}
\end{figure}

In the function \( f^{(\text{M})} \), the reference saliency point \( p^{(\text{ref})}_i\) is compared against all answer saliency points in the list \( P^{(\text{ans})} \) to evaluate the semantic matching. If the \(i\)-th reference saliency point \( p^{(\text{ref})}_i\) finds the best match from \( P^{(\text{ans})} \), the output \( a_i=j\), where \(j\) is the index of the best-matched answer saliency point.  If no match is found, the output is \( a_i=-1\). As illustrated in Figure \ref{matcher_scorer}, there are three matched pair of saliency points, therefore, the output of matching stage for this example is \([4,2,3,-1]\).

To achieve this, an LLM-based matcher is proposed, leveraging the chain-of-thought (CoT) technique to guide the LLM to evaluate each answer saliency point in  \( P^{(\text{ans})} \) sequentially and to identify the best match (if one exists). Moreover, the prompt for matcher is highly flexible, allowing for specific instructions depending on the task requirements, \textit{e.g.}, a good answer should focus more on numerical details such as the revenue or cost during the matching stage. Saliency point matching prompts used by LLM are listed in Appendix~\ref{sec:match_prompt}.

\subsection{Stage 3: Scoring} \label{sec_scoring}
A scoring function \(f^{(\text{S})}\) is introduced in the last stage of \ems evaluation pipeline. Given \( p^{(\text{ref})}_i\) and matched answer point \( p^{(\text{ans})}_{a_i} \), the alignment score is computed as
\begin{equation}
    s^{(\text{ref})}_i =
    \begin{cases}
    f^{(\text{S})}\left(p^{(\text{ref})}_i,p^{(\text{ans})}_{a_i}\right) & a_i\neq-1, \\
    0 & a_i=-1,
    \end{cases}
    \label{equ_scoring}
\end{equation}
where \(s^{(\text{ref})}_i\in[0,1]\) and \(i=1,\dots,N\).

An LLM is employed for this stage, which estimates the score for a pair of saliency points based on the extent to which the details of two saliency points are aligned. As in the previous stages, the LLM provides granular flexibility to accommodate task-specific criteria in addition to the general ones, which are defined in our pipeline as follows:
\begin{itemize}
    \item \textbf{Completeness}: Does the candidate answer point capture all the essential aspects of the reference point?
    \item \textbf{Accuracy}: Is the information in the candidate answer point contextually correct?
    \item \textbf{Level of Detail}: Does the candidate answer point provide sufficient details, matching the depth of the reference point?
\end{itemize}

While the matching stage primarily focuses on ranking and assigning binary scores, \textit{i.e.}, 1 for matched pairs and 0 for unmatched pairs, the scoring stage differs by assigning a soft score (continuous from 0 to 1) to each matched pair of saliency points. With soft scores, the detailed level of semantic alignment between two saliency points can be accurately reflected in the metrics. It is more useful than a binary score as a reference for fine-tuning the LLM or adjusting the prompt to have more accurate LLM-generated answers. The details scoring prompt is in Appendix~\ref{sec:score_prompt}.

Although the matching and scoring stages can be handled by a single LLM for both tasks, using one LLM for a combined and complex task may result in poorer performance and higher costs compared to using multiple LLMs, each handling a simpler sub-task \citep{yue2024large}. Furthermore, having a separate scoring stage allows the pipeline to integrate with conventional NLP metrics. For example, if ROUGE is chosen as the scoring mechanism, output \(s^{(\text{ref})}_i\) becomes the ROUGE score between two saliency points. Since long-form answers are decomposed into saliency points, which are typically just one or two sentences long, conventional metrics perform better in this scenario than when comparing long-form answers directly.

\subsection{EMS Metrics} \label{sec_metric}
After the Extract-Match-Score pipeline, the long-form answer and reference are converted to a matching list with the corresponding alignment scores assigned for the matched pairs. In our \ems pipeline, EMS-Recall, EMS-Precision, and EMS-F1 are computed as the metrics.

\textbf{EMS-Recall} is the ``proportion'' of the long-form reference which are accurately captured in the semantic level, which is computed by 
\begin{equation}
    \text{EMS-Recall} = \frac{\sum_{i=1}^{N}s^{(\text{ref})}_i}{N}
    \label{equ_ems_rec}
\end{equation}
As \(s^{(\text{ref})}_i\) is the soft score, EMS-Recall represents the recall calculated over the level of detailed information, instead of a simple recall over the sentences or claims.

\textbf{EMS-Precision} is the ``proportion'' of the long-form answer that is aligned with the reference. To calculate EMS-Precision, the alignment score for reference saliency points \(s^{(\text{ref})}_i\) are first mapped to answer saliency points by 
\begin{equation}
    s^{(\text{ans})}_j=
    \begin{cases}
    \max\left(\{s^{(\text{ref})}_i|a_i=j\}\right) & j\in A, \\
    0 & j\notin A.
    \end{cases}    
    \label{equ_mapping}
\end{equation}
It is important to note that in the implementation, multiple reference saliency points may match to one single answer saliency point. It may be caused by the repetition of similar information within the long-form reference. In this case, the maximum score is selected to represent the alignment of this answer saliency point. Subsequently, EMS-Precision is computed as 
\begin{equation}
    \text{EMS-Precision} = \frac{\sum_{j=1}^{M}s^{(\text{ans})}_j}{M}
    \label{equ_ems_prec}
\end{equation}
Similarly, EMS-Precision refers to information-level precision of the LLM-generated long-form answer. 

\textbf{EMS-F1} is computed by 
\begin{equation}
    \text{EMS-F1}=\frac{2\times\text{EMS-Precision}\times\text{EMS-Recall}}{\text{EMS-Precision}+\text{EMS-Recall}}.
    \label{equ_ems_f1}
\end{equation}

As discussed in Section \ref{sec_scoring}, the score \(s_i\) can be derived from either LLM-based scorer or conventional NLP metrics. With the ROUGE score example, ROUGE-F1 is employed for saliency point-level scoring. Subsequently, EMS-F1 is calculated to evaluate the entire long-form answer. As several scorers are implemented and tested, to distinguish pipelines between different scorers, in the following sections:
\begin{itemize}
    \item \( \text{EMS}^{\text{(LLM)}} \) denotes the EMS evaluation pipeline with LLM-based scorer;
    \item \( \text{EMS}^{\text{(ROUGE)}} \) denotes the pipeline where the scorer is implemented by ROUGE-F1;
    \item \( \text{EMS}^{\text{(BERTScore)}} \) denotes the pipeline where the scorer is implemented by BERT score.
\end{itemize}
The long-form answer level recall, precision, and F1 are then calculated with the outputs from different scorers, \textit{e.g.}, \( \text{EMS}^{\text{(LLM)}}\text{-F1} \).

\section{Experiment Results}
\subsection{Dataset}
We have constructed a financial Document-Based QA dataset specifically focusing on long-context and long-form answers. This dataset uses the earnings call transcripts from the third quarter of 2024 for the 10 largest S\&P 500 constituents at the time of writing this paper, namely, \textit{Apple}, \textit{NVIDIA}, \textit{Microsoft}, \textit{Alphabet}, \textit{Amazon}, \textit{Meta Platforms}, \textit{Tesla}, \textit{Broadcom}, \textit{JPMorgan Chase}, and \textit{Eli Lilly}. The dataset captures a rich set of financial discussions, encompassing detailed business strategies, performance analyses, and forward-looking statements, which is a useful resource for financial NLP research and applications.

We analyze the Q\&A sections of several earnings call transcripts to identify common themes and question formats used by analysts. We then use these insights to design five questions that cover commonly referred topics. The questions are further refined iteratively with the assistance of LLMs in order to produce high-quality, contextually relevant answers that reflect real-world financial analysis. Each question is designed to address specific aspects of company performance, such as revenue growth, operational challenges, strategic initiatives, financial forecasts, and competitive positioning. The five questions are listed in Appendix~\ref{five_questions}.

To construct high-quality answers that serve as a reference for the evaluation of other LLM-generated answers, we leverage the state-of-the-art LLMs (GPT-4o and Mistral Large). We prompt each LLM to answer each question individually and then generate a consolidated answer by instructing GPT-4o to combine and refine the answers, taking into account their strengths and resolving any discrepancy by referring to the original document. The complete prompts are shown in Appendix~\ref{Answer_Generation_Prompts}.

\subsection{Baseline Methods}
We consider both classical numerical metrics and more recent LLM-based evaluation frameworks.
\begin{itemize}
    \item \textbf{BLEU:} BLEU~\citep{papineni2002bleu} measures the match between a candidate answer and a reference answer by looking at the percentage of overlapping n-grams. 
    \item \textbf{ROUGE:} While BLEU focuses on precision, measuring how many n-grams in the candidate answer appear in the reference answer, ROUGE~\citep{lin2004rouge} emphasizes recall, evaluating how much of the reference answer is captured by the candidate.
    \item \textbf{BERTScore:} BERTScore~\citep{Zhang*2020BERTScore} uses the pre-trained contextualized token embeddings from BERT to calculate the similarity between candidate answer and reference sentences.
    \item \textbf{RAGChecker:} RAGChecker~\citep{ru2024ragchecker} is a latest LLM-based evaluation framework. Although designed for RAG systems, it can assess end-to-end performance via its precision and recall metrics by extracting claims from the model responses and then verifying their entailment against the provided ground truth.
\end{itemize}

\subsection{Results}
\label{sec:results}

We have shown the evaluation results averaged across ten companies and all the questions in Table~\ref{main_results_2}. Three LLM models are selected from the Llama3.2 family with the number of model parameters ranging from 1 billion to 90 billion. By this means, we avoid the performance nuances brought by different LLM variations and focus on validating the scaling effect of LLMs with different evaluation metrics. 

Table~\ref{main_results_2} presents that both RAGChecker and our proposed EMS metrics show more nuanced assessments of the response quality. We could clearly see the upward trend of the evaluation figures as the LLM model size increases from 1 bilion to 90 billion. In contrast, conventional evaluation metrics such as BLEU, ROUGE and BERTScore fail to capture this trend with different model sizes. Based on the well-established understanding that larger models offer superior performance, our proposed EMS evaluation is more effective in long-form financial analyses.

\begin{table}
\centering
\caption{Evaluation Results of Ten Companies over All Questions }
\label{main_results_2}
\begin{tblr}{
  cell{1}{1} = {c=2}{},
  cell{2}{1} = {c=2}{},
  cell{2}{3} = {c},
  cell{2}{4} = {c},
  cell{2}{5} = {c},
  cell{3}{1} = {r=3}{},
  cell{3}{3} = {c},
  cell{3}{4} = {c},
  cell{3}{5} = {c},
  cell{4}{3} = {c},
  cell{4}{4} = {c},
  cell{4}{5} = {c},
  cell{5}{3} = {c},
  cell{5}{4} = {c},
  cell{5}{5} = {c},
  cell{6}{1} = {c=2}{},
  cell{6}{3} = {c},
  cell{6}{4} = {c},
  cell{6}{5} = {c},
  cell{7}{1} = {r=3}{},
  cell{7}{3} = {c},
  cell{7}{4} = {c},
  cell{7}{5} = {c},
  cell{8}{3} = {c},
  cell{8}{4} = {c},
  cell{8}{5} = {c},
  cell{9}{3} = {c},
  cell{9}{4} = {c},
  cell{9}{5} = {c},
  cell{10}{1} = {r=3}{},
  cell{10}{3} = {c},
  cell{10}{4} = {c},
  cell{10}{5} = {c},
  cell{11}{3} = {c},
  cell{11}{4} = {c},
  cell{11}{5} = {c},
  cell{12}{3} = {c},
  cell{12}{4} = {c},
  cell{12}{5} = {c},
  cell{13}{1} = {r=3}{},
  cell{13}{3} = {c},
  cell{13}{4} = {c},
  cell{13}{5} = {c},
  cell{14}{3} = {c},
  cell{14}{4} = {c},
  cell{14}{5} = {c},
  cell{15}{3} = {c},
  cell{15}{4} = {c},
  cell{15}{5} = {c},
  cell{16}{1} = {r=3}{},
  cell{16}{3} = {c},
  cell{16}{4} = {c},
  cell{16}{5} = {c},
  cell{17}{3} = {c},
  cell{17}{4} = {c},
  cell{17}{5} = {c},
  cell{18}{3} = {c},
  cell{18}{4} = {c},
  cell{18}{5} = {c},
}
\hline
Metric/LLM                       &           & {llama3.2\_1b\_\\instruct} & {llama3.2\_11b\_\\vision\_instruct} & {llama3.2\_90b\_\\vision\_instruct} \\
\hline
BLEU                             &           & 0.02                             & 0.04                                & 0.03                                \\
\hline
ROUGE                            & Precision & 0.23                             & 0.21                                & 0.22                                \\
                                 & Recall    & 0.14                             & 0.21                                & 0.19                                \\
                                 & F1        & 0.17                             & 0.20                                & 0.20                                \\
\hline
BERTScore                        &           & 0.84                             & 0.85                                & 0.85                                \\
\hline
RAGChecker                       & Precision & 0.45                             & 0.63                                & 0.63                                \\
                                 & Recall    & 0.15                             & 0.31                                & 0.35                                \\
                                 & F1        & 0.21                             & 0.40                                & 0.44                                \\
\hline
\textbf{EMS\textsuperscript{(ROUGE)}}     & \textbf{Precision} & 0.08                             & 0.17                                & 0.17                                \\
                                 & \textbf{Recall}    & 0.07                             & 0.12                                & 0.13                                \\
                                 & \textbf{F1}        & 0.07                             & 0.13                                & 0.15                                \\
\hline
\textbf{EMS\textsuperscript{(BERTScore)}} & \textbf{Precision} & 0.38                             & 0.63                                & 0.65                                \\
                                 & \textbf{Recall}    & 0.34                             & 0.42                                & 0.47                                \\
                                 & \textbf{F1}        & 0.35                             & 0.49                                & 0.54                                \\
\hline
\textbf{EMS\textsuperscript{(LLM)}}       & \textbf{Precision} & 0.23                             & 0.45                                & 0.47                                \\
                                 & \textbf{Recall}    & 0.21                             & 0.30                                & 0.36                                \\
                                 & \textbf{F1}        & 0.21                             & 0.35                                & 0.40                                \\
\hline
\end{tblr}
\end{table}

\subsection{Discussion}
As mentioned in Section~\ref{sec:results}, compared to conventional metrics such as BLEU, ROUGE and BERTScore, both our EMS approach and RAGChecker offer significantly more nuanced assessments of answer quality. BLEU and ROUGE compute n-grams level matching, which may not capture the overall meaning or the entire context of the response. In addition, the F1 scores produced by RAGChecker and our method \( \text{EMS}^{\text{(LLM)}} \) both range between 0.2 and 0.5, indicating that a naive question-answer approach for long triplets in financial analysis may not yield satisfactory results. Practitioners should therefore expect to invest substantial effort into refining their solutions. Our proposed EMS evaluation metric, at the same time, could be helpful in evaluating different solutions effectively, aiding practitioners in selecting suitable backbones for financial analysis.

Compared to the competing  approach of RAGChecker, our proposed method offers three key advantages. First, EMS extracts the saliency points, which are carefully designed to maintain all the details from the original long-form answer and could provide practitioners tangible insights on the overall financial analysis.

Second, our extensible EMS method opens the door to multiple scoring mechanisms, including conventional methods, such as BLEU and ROUGE, non-LLMs model-based scoring like BERTScore, and LLM-based scoring. This flexibility allows practitioners to incorporate the advantages of different evaluation metrics and to tailor their scoring strategy based on specific requirements and computation resources. In contrast, RAGChecker is constrained by its configuration reliance on a LLM to generate scores. 

Third, rather than the binary (hit-or-miss) outcome used by RAGChecker, our approach employs a continuous scale (0--1) scoring mechanism, enabling finer-grained distinctions in answer quality. This is particularly valuable in pinpointing specific areas of improvement for long-form financial analyses, offering practitioners clearer guidance on how to enhance system performance. 

\section{Conclusion}
In this study, we first highlight the limitations of conventional evaluation approaches in distinguishing generation quality for financial long triplets. To address this issue, we propose an extensible framework, EMS, that assesses answer quality by extracting saliency points, matching with answer saliency points, and scoring the alignment. We conducted experiments on a self-constructed financial Document-Based QA dataset and provided detailed analysis on EMS over both conventional and latest approaches. We envisage that  EMS could aid practitioners in answer assessment within the context of long triplets, and facilitate researchers to explore this complex yet essential scenario beyond financial NLP.

\subsubsection*{Limitation}
This study has three main limitations. First, we primarily evaluated the inference capabilities of LLMs on qualitative information of earning call transcripts and did not dive into the mathematical reasoning, which is crucial in financial NLP  \citep{chen-etal-2021-finqa,chen-etal-2022-convfinqa,krumdick-etal-2024-bizbench,srivastava-etal-2024-evaluating}. Second, LLMs were employed as the primary annotator to generate silver answers, which may introduce biases and errors compared to panel discussions by human experts \citep{felkner-etal-2024-gpt,ronningstad-etal-2024-gpt}. We view our effort as an initial study and future studies should recruit financial professionals to develop the golden answer and validate the effectiveness of the proposed EMS evaluation. Third, the current evaluation mainly focused on the schematic comparison and future work will develop a more systematic \citep{potluri-etal-2023-concise,koh-etal-2022-far} and multifaceted approach to consider additional factors such as logical consistency and factuality \citep{cho-etal-2019-towards,min-etal-2023-factscore,song-etal-2024-veriscore}.

\subsubsection*{Ethics Statement}
To facilitate reproducibility and ensure the consistency of our findings, we have included the detailed prompt in Appendix. The original earnings call transcripts are publicly available. We acknowledge the potential for geographic bias in our dataset, as all samples were derived from constituents of the S\&P index. Publicly accessible language models were employed at various stages of our experiments. Given their inherent stochasticity, these models may generate unexpected outputs. To mitigate risks to model safety, we refrained from fine-tuning the models.

\subsubsection*{Disclaimer}
This paper is intended solely for informational purposes and the technique is not a business practice of American Express. The earnings call transcripts is used for demonstration of our evaluation framework and analysis on them is not intended to constitute investment research, advice, recommendations, or an offer to buy or sell securities. The opinions, findings and conclusions of this paper are those of the authors alone and do not reflect the views of  American Express.

\bibliography{iclr2025_conference}
\bibliographystyle{iclr2025_conference}

\appendix
\section{Appendix}
\subsection{Five Questions Designed for the Financial QA Dataset} \label{five_questions}
\subsubsection*{Question 1}
Can you provide a detailed analysis of the key themes, topics, and forward-looking guidance presented during the earnings call? Please include specific figures, strategic initiatives, and performance metrics highlighted by the executives, as well as any macroeconomic factors that may influence the business. Additionally, how do these elements shape the company's future trajectory and overall market position?
\subsubsection*{Question 2}
In the context of an earnings call, could you outline the specific risks and uncertainties that management highlights regarding future performance? Please include insights into challenges related to investments, cybersecurity, and operational strategies, as well as any external factors that could significantly impact the company's operations and financial outlook. Additionally, please reference relevant information on risk factors from the company's recent filings or official communications and discuss how these challenges might affect the company's financial performance and future growth.
\subsubsection*{Question 3}
To gain insights into the company’s strategic direction and priorities, what key elements and initiatives highlighted in the earnings call transcript reflect their approach to growth, innovation, and market positioning? Please elaborate on specific strategies related to product categories, research and development, capital expenditures, partnerships, and how these initiatives aim to enhance customer value and operational efficiency, ultimately influencing the company's trajectory moving forward.
\subsubsection*{Question 4}
In the context of an earnings call transcript, what significant industry trends and current macroeconomic conditions are addressed by company executives that may affect both the company's performance and the broader industry landscape? Please highlight insights related to competitive dynamics, commodity markets, and specific challenges faced by the company, including relevant data points and strategic considerations that underscore the influence of these trends and conditions.
\subsubsection*{Question 5}
In the context of an earnings call, how can we gain insights into the company's strategies and priorities regarding cash flow management? Specifically, what key components should we examine, such as cash flow from operations, capital expenditures, and free cash flow trends? Additionally, what factors contribute to the resilience and growth of free cash flow margins, and how does the organization balance investment opportunities with shareholder returns amidst market fluctuations? What does this reveal about the company's approach to financial performance and resource allocation for future growth initiatives?

\subsection{Answer Generation Prompts}
\label{Answer_Generation_Prompts}

\begin{tcolorbox}[colback=white, colframe=black, title=Individual LLM Answer Generation Prompt]
Instructions: \\
1. Review the Question: Carefully read the provided question and ensure that you fully understand what is being asked.\\

2. Make sure that the answer is of high quality with the following consideration:\\
- Clarity: Is the answer easy to understand? Are any terms or concepts unclear?\\
- Completeness: Does the answer fully address the question, or are there missing aspects that need to be covered?\\
- Accuracy: Were there factual errors or outdated information in the previous answer? Is the answer well-supported by evidence?\\
- Conciseness: Is the answer concise without sacrificing important details? Avoid unnecessary verbosity.\\

3. Final Review: Once the draft of the answer is complete, re-read the answer to ensure that:\\
- All aspects of the question are addressed.\\
- The answer is factually accurate, clear, and well-supported by evidence. Don’t include any citation in the final answer.\\

Question:\\
$<$Question$>$

\end{tcolorbox}
\noindent\begin{minipage}{\textwidth}
\captionof{figure}{The prompt used to generate the individual answers from GPT-4o and Mistral Large.}\label{box}
\end{minipage}

\begin{tcolorbox}[breakable, colback=white, colframe=black, title=Consolidated Answer Generation Prompt]
\textbf{Instructions:}\\

You will be provided with:\\

1. An earnings call transcript.\\
2. The original question asked.\\
3. Three versions of answers generated by different LLMs.\\

Your task is to combine these answers into a single, high-quality final response. Follow the instructions below carefully:\\

1. \textbf{Review the Original Question:}
\begin{tabbing}
    \hspace{0.3cm}  \= Carefully read the provided question to ensure you fully understand its requirements.\\
\end{tabbing}
2. \textbf{Analyze the Provided Answers:}
\begin{tabbing}
    \hspace{0.3cm}  \= - Identify the strengths of each answer. Consider clarity, completeness, accuracy, and\\ \hspace{0.6cm}conciseness.\\
    \hspace{0.3cm}  \= - Look for overlap, unique points, and discrepancies among the answers.
\end{tabbing}
3. \textbf{Combine and Refine:}
\begin{tabbing}
    \hspace{0.3cm}  \= - Synthesize the best elements of all three answers into a cohesive, well-structured\\
    \hspace{0.6cm}response.\\
    \hspace{0.3cm}  \= - Ensure the final answer is clear, concise, and comprehensive, addressing all aspects of\\ \hspace{0.6cm}the original question.\\
    \hspace{0.3cm}  \= - Resolve any discrepancies or conflicting information by considering the context of the\\
    \hspace{0.6cm}earnings call transcript.
\end{tabbing}
4. \textbf{Final Quality Check:}
\begin{tabbing}
    \hspace{0.3cm}  \= - Ensure the final response is factually accurate, clear, and free of redundancy.\\
    \hspace{0.3cm}  \= - Avoid unnecessary verbosity while retaining all essential details.\\
    \hspace{0.3cm}  \= - Do not include citations or direct references to the original sources.
\end{tabbing}

\textbf{Output Format:}\\
Provide a single, polished answer that effectively combines the key points, figures, strategic insights, and forward-looking guidance from the provided answers. Ensure it fully addresses the original question.\\

\textbf{Original Question:}\\
$<$Question$>$\\

\textbf{Earnings Call Transcript (for reference):}\\
$<$Attached pdf file$>$\\

\textbf{Answer Versions to Combine:}\\
Answer Version 1: $<$Answer\_Version\_1$>$\\
Answer Version 2: $<$Answer\_Version\_2$>$
\end{tcolorbox}
\noindent\begin{minipage}{\textwidth}
\captionof{figure}{The prompt used to form the final answer by combining the answers from GPT-4o and Mistral Large.}\label{box}
\end{minipage}

\subsection{Saliency Points Extraction Prompt}
\label{sec:extract_prompt}
\begin{tcolorbox}[breakable, colback=white, colframe=black, title=Saliency Points Extraction Prompt]
Task Description:\\
You are tasked with extracting bullet points (key points) from the provided Candidate Answer.\\

Guidelines:\\

Extraction Task: Focus on extracting the content without evaluating its factual accuracy.\\
Conciseness: Each bullet point must be no longer than two sentences.\\
Include Overlap and Repetition: Ensure that the extracted key points retain all overlapping and repeated information as presented in the original content. Do not remove or merge redundant details; if similar information appears multiple times, include it in the extracted key points as is.\\
Format: The output should be in a List of Strings, where each string is a distinct bullet point.\\

In-Context Examples:\\

Example 1\\
Candidate Answer:\\
Tesla Bot, also known as Optimus, is a humanoid robot developed by Tesla Inc. It is designed to perform repetitive or unsafe tasks, leveraging Tesla's advancements in AI and robotics. Standing approximately 5'8" tall and weighing around 125 pounds, Optimus features human-like proportions and a sleek design. It uses Tesla's AI technology, including computer vision and self-learning algorithms, to navigate and interact with its environment. The bot is envisioned as a tool to enhance productivity and safety in industrial, household, and other labor-intensive settings.\\

Bullet Points:\\
$[$\\
"Tesla Bot is a humanoid robot developed by Tesla Inc.",\\
"Tesla Bot is also known as Optimus.",\\
"Tesla Bot is designed to perform repetitive or unsafe tasks in industrial, household, and other labor-intensive settings.",\\
"Tesla Bot is approximately 5'8" tall and weighs around 125 pounds, with human-like proportions and a sleek design.",\\
"Tesla Bot uses AI technology to navigate and interact with its environment."\\
$]$\\

Example 2\\
Candidate Answer:\\
Lenovo is a global technology company headquartered in Beijing, China, and Morrisville, North Carolina, USA. Founded in 1984, it is renowned for designing, manufacturing, and selling computers, smartphones, servers, and other technology products. Lenovo is a market leader in PCs, offering popular ThinkPad and Yoga series laptops, alongside gaming-focused Legion products. Lenovo is a market leader in PCs, offering popular ThinkPad and Yoga series laptops, alongside gaming-focused Legion products.\\

Bullet Points:\\
$[$\\
"Lenovo is a global technology company with headquarters in Beijing, China, and Morrisville, North Carolina, USA.",\\
"Founded in 1984, Lenovo specializes in designing, manufacturing, and selling computers, smartphones, servers, and other tech products.",\\
"Lenovo is a PC market leader, known for ThinkPad and Yoga laptops, as well as gaming-focused Legion products.",\\
"Lenovo is a PC market leader, known for ThinkPad and Yoga laptops, as well as gaming-focused Legion products."\\
$]$\\

Your Task:\\
Generate the bullet points for the following candidate answer based on the guidelines above.\\

Candidate Answer:\\
$<$ans$>$\\

Bullet Points:
\end{tcolorbox}
\noindent\begin{minipage}{\textwidth}
\captionof{figure}{The prompt used to extract saliency points from candidate answers.}\label{extract_prompt}
\end{minipage}

\subsection{Salient Points Matching Prompt}
\label{sec:match_prompt}
\begin{tcolorbox}[breakable, colback=white, colframe=black, title=Saliency Points Matching Prompt]
Task Description:\\
Your task is to identify the most matched keypoint in the provided Candidate Keypoint List for a given Reference Keypoint. Each item in the candidate keypoint list contains one index and one string representin the keypoint.\\

Matching Criteria:\\

The matched keypoint should represent the same information as the reference keypoint.
If no keypoint in the candidate list matches the reference, return -1.\\

Pay special attention to matching numerical details between the two key points. If both key points mention the same specific numbers (e.g., "growth of 30\%"), this significantly increases their likelihood of being a match. However, ensure the context of the numbers aligns. For example, "35\% growth" should not be matched to "35 people" as the numbers pertain to entirely different contexts. Always evaluate numerical information within the broader context of the key points.\\

Input Format:\\

Reference Keypoint: "..."\\
Candidate Keypoint List: $[$ 0: "...", 1: "...", 2: "..." $]$\\
Output Format:\\

The output should be a single integer index of the best-matched candidate keypoint.
If no match is found, output -1.
Do not include any explanations or reasoning in the output.\\
In-Context Examples:\\

Example 1\\
Reference Keypoint:\\
"Lenovo is a global technology company with headquarters in different countries."\\
Candidate Keypoint List:\\
$[$\\
"Lenovo is a global technology company with headquarters in Beijing, China, and Morrisville, North Carolina, USA.",\\
"Founded in 1984, Lenovo specializes in designing, manufacturing, and selling computers, smartphones, servers, and other tech products.",\\
"Lenovo is a PC market leader, known for ThinkPad and Yoga laptops, as well as gaming-focused Legion products."\\
$]$\\
Matched Index:\\
0\\

Example 2\\
Reference Keypoint:\\
"The alias of Tesla Bot is Optimus."\\
Candidate Keypoint List:\\
$[$\\
"Tesla Bot is a humanoid robot developed by Tesla Inc.",\\
"Tesla Bot is also known as Optimus.",\\
"Tesla Bot is designed to perform repetitive or unsafe tasks in industrial, household, and other labor-intensive settings.",\\
"Tesla Bot is approximately 5'8" tall and weighs around 125 pounds, with human-like proportions and a sleek design.",\\
"Tesla Bot uses AI technology to navigate and interact with its environment."\\
$]$\\
Matched Index:\\
1\\

Example 3\\
Reference Keypoint:\\
"PCIe bifurcation allows a single PCIe slot to be divided into multiple lanes."\\
Candidate Keypoint List:\\
$[$\\
"Flexible Leasing Options: Offers a variety of leasing solutions, including wet lease, dry lease, and lease-purchase agreements tailored to meet airline requirements.",\\
"Comprehensive Fleet Management: Provides access to a wide range of aircraft models, ensuring compatibility with operational needs and passenger demands.",\\
"Cost-Effective Solutions: Reduces the financial burden of aircraft ownership through competitive lease terms and efficient asset utilization.",\\
"Global Network: Connects airlines with a diverse pool of lessors and aircraft owners across the world."\\
$]$\\
Matched Index:\\
-1\\

Your Task:\\
Find the best matched keypoint index from the candidate keypoint list for the given reference keypoint.\\

Reference Keypoint:\\
$<$ref$>$\\

Candidate Keypoint List:\\
$<$candid$>$\\

Matched Index:

\end{tcolorbox}
\noindent\begin{minipage}{\textwidth}
\captionof{figure}{The prompt used to match candidate saliency points with reference saliency points.}\label{match_prompt}
\end{minipage}

\subsection{Alignment Scoring Prompt}
\label{sec:score_prompt}
\begin{tcolorbox}[breakable, colback=white, colframe=black, title=Alignment Scoring Prompt]
Task Description:\\
Evaluate the similarity between two key points provided by a language model and assign a score between 0 and $\{$max\_score$\}$ based on the following criteria:\\

Score $\{$max\_score$\}$: If the two key points are exactly matched, including all details.\\

Intermediate Score: If the two key points share some information but are not fully matched. The more they match, the closer the score should be to $\{$max\_score$\}$.\\

Score 0: If the two key points are entirely irrelevant to each other.\\

Guidelines:\\

The higher the similarity, the higher the score (closer to $\{$max\_score$\}$).
The lower the similarity, the lower the score (closer to 0).
Output must be a single integer from 0 to $\{$max\_score$\}$.
Do not provide any explanations or reasoning in your response.\\

In-Context Examples:\\
Example 1:\\
Input:\\
Keypoint 1: The Gemini model supports over 2 billion monthly users across products like Search, Google Cloud, YouTube, and Google Maps, with API calls increasing 14x in six months.\\
Keypoint 2: Sundar Pichai mentions that Gemini is now available on GitHub Copilot, with over 2 billion monthly users across all seven products.\\

Output: 7\\

Example 2:\\
Input:\\
Keypoint 1: Google Cloud revenue increased 35\% YoY to \$11.4 billion, with a 17\% operating margin, driven by AI solutions like Vertex AI and BigQuery.\\
Keypoint 2: Alphabet's Cloud revenue grew 35\% year-over-year, with operating margins increasing to 17\%.\\

Output: 10\\

Example 3:\\
Input:\\
Keypoint 1: AI-driven features like AI Overviews and Google Lens are transforming user experiences and increasing engagement.\\
Keypoint 2: The executives discuss the benefits of GenAI, including reduced costs, greater customer engagement, and faster response times.\\

Output: 5\\

Example 4:\\
Input:\\
Keypoint 1: Alphabet plans to advance its AI portfolio with the next-generation Gemini model and broader enterprise integrations.\\
Keypoint 2: The executives discuss various AI-powered products and services, including Gemini, Google Cloud AI, and Google DeepMind.\\

Output: 1\\

Input:\\
Keypoint 1:\\
$<$kp1$>$\\

Keypoint 2:\\
$<$kp2$>$\\

Output:\\
Matching Score:

\end{tcolorbox}
\noindent\begin{minipage}{\textwidth}
\captionof{figure}{The prompt used to determine the alignment score between two saliency points.}\label{score_prompt}
\end{minipage}
\end{document}

%% file: math_commands.tex
\usepackage{amsmath,amsfonts,bm}

\def\eqref#1{equation~\ref{#1}}

\def\1{\bm{1}}

\DeclareMathAlphabet{\mathsfit}{\encodingdefault}{\sfdefault}{m}{sl}
\SetMathAlphabet{\mathsfit}{bold}{\encodingdefault}{\sfdefault}{bx}{n}

